%% file: main.tex
\documentclass[11pt]{article}

\usepackage[most]{tcolorbox}
\usepackage[preprint]{acl}
\usepackage{amsmath}
\usepackage{times}
\usepackage{xcolor}

\usepackage{latexsym}
\usepackage{booktabs}
\usepackage{graphicx}
\usepackage[table]{xcolor}
\usepackage{multirow}
\usepackage{amssymb}
\usepackage{amsmath}
\usepackage[T1]{fontenc}

\usepackage[utf8]{inputenc}

\usepackage{microtype}

\usepackage{inconsolata}

\usepackage{graphicx}

\title{ReCo: Reweighting GRPO Against Distributional Concentration}

\author{
  \textbf{Junoh Park}\thanks{Equal contribution.}
  \quad
  \textbf{Junseo Hwang}\footnotemark[1]
  \quad
  \textbf{Wonguk Cho}
  \quad
  \textbf{Taesup Kim}\thanks{Corresponding author.}
  \\
  Graduate School of Data Science, Seoul National University
  \\
  \texttt{\{wnsdh0418, hjunseoh, wongukcho, taesup.kim\}@snu.ac.kr}
}

\begin{document}
\maketitle
\input{Sec/abstract}
\input{Sec/Introduction}
\input{Sec/Preliminaries}
\input{Sec/Method}

\input{Sec/Experiments}
\input{Sec/Related_work}
\input{Sec/Conclusion}

\section*{Limitations and Discussion}
\label{sec:discussion_limitations}

This work focuses on RLVR with sparse binary outcome rewards, where feedback is given mainly by final-answer correctness. While this setting is standard for mathematical reasoning, many tasks provide denser or less discrete supervision, such as feedback on intermediate reasoning steps, partial progress, or subjective output quality. Since such rewards may be continuous or multi-dimensional, extending ReCo beyond sparse binary rewards remains an important direction for future work.

ReCo also applies response-level and token-level reweighting uniformly during training. A more adaptive variant could decide when and how strongly to apply each correction based on the rollout distribution or training dynamics. Moreover, ReCo reweights gradients after rollouts are sampled from the policy, but does not directly change the sampling process itself. Combining ReCo with exploration-oriented decoding strategies may further improve rollout diversity and broaden the set of reasoning paths used for learning.

\section*{Ethics Statement}
\label{sec:ethics}
This work proposes ReCo, a reweighting method for reinforcement learning with verifiable rewards, and does not involve human subjects, crowdsourcing, or the collection of new data. All experiments use publicly available mathematical and code reasoning benchmarks and open-source pretrained language models, and we do not introduce any new datasets containing personal or sensitive information. We do not anticipate direct negative societal impacts from this work. As with any method that improves the reasoning capability of language models, however, downstream applications should remain subject to appropriate oversight to mitigate potential misuse. We report all hyperparameters and experimental settings needed to reproduce our results in the Appendix.

\bibliography{main}

\clearpage
\appendix
\input{Sec/Appendix}








\end{document}

%% file: Sec/Abstract.tex
\begin{abstract}
Group Relative Policy Optimization (GRPO) has become a standard reinforcement learning method for post-training language models. Recent work shows that GRPO can reduce the base model's reasoning capacity and underperform it in Pass@$k$ when $k$ is large, indicating reduced coverage of reasoning paths. We find that this reduction is associated with GRPO concentrating on responses that the base model already generates with high probability. We trace this concentration to two mechanisms in the GRPO update. At the response level, high-probability responses dominate the group gradient through repeated occurrence. At the token level, GRPO's importance ratio scales gradients, further reinforcing tokens that become more likely under the current policy. We propose \textbf{ReCo}, a reweighting method that addresses both effects. Response contributions are normalized by their expected occurrence within the rollout group, and the token-level importance ratio is replaced with a variance-based ratio that gives larger update scale to non-saturated decision points where alternative token choices remain plausible. Across Qwen2.5-Math-1.5B/7B and Llama-3.1-8B-Instruct on five mathematical reasoning benchmarks, ReCo improves Pass@$k$ for large values of $k$ and is comparable to GRPO for small values of $k$.
\end{abstract}


%% file: Sec/Introduction.tex
\section{Introduction}
\label{sec:intro}
 
Reinforcement Learning with Verifiable Rewards (RLVR) has become a standard post-training approach for improving LLM reasoning, especially in domains such as mathematics and coding where correctness can be automatically verified~\citep{jaech2024openai,guo2025deepseekr1,yu2025dapo,zhang2025srpo}. Among RLVR methods, Group Relative Policy Optimization (GRPO) has been widely adopted for training reasoning models. It samples a group of rollouts for each prompt and computes advantages relative to the group mean, providing stable policy gradients without a critic~\citep{shao2024deepseekmath}.

Despite its empirical success, GRPO has key limitations. While GRPO improves sampling efficiency, it has been observed to reduce the reasoning capacity and the diversity of problem-solving trajectories that the base model originally possessed~\citep{yue2025does,
wu2026invisibleleashrlvrescape,li2025divergence}. 
This limitation is often observed through Pass@$k$, where GRPO outperforms the base model for small values of $k$ but fails to surpass or even falls below it for large values of $k$~\citep{yue2025does,zhu202nsr,liu2025prorl}. While this pattern has been widely reported, it remains unclear what distributional change under GRPO training leads to this limitation.
 
\begin{figure}[t]
    \centering
    \includegraphics[width=1\linewidth]{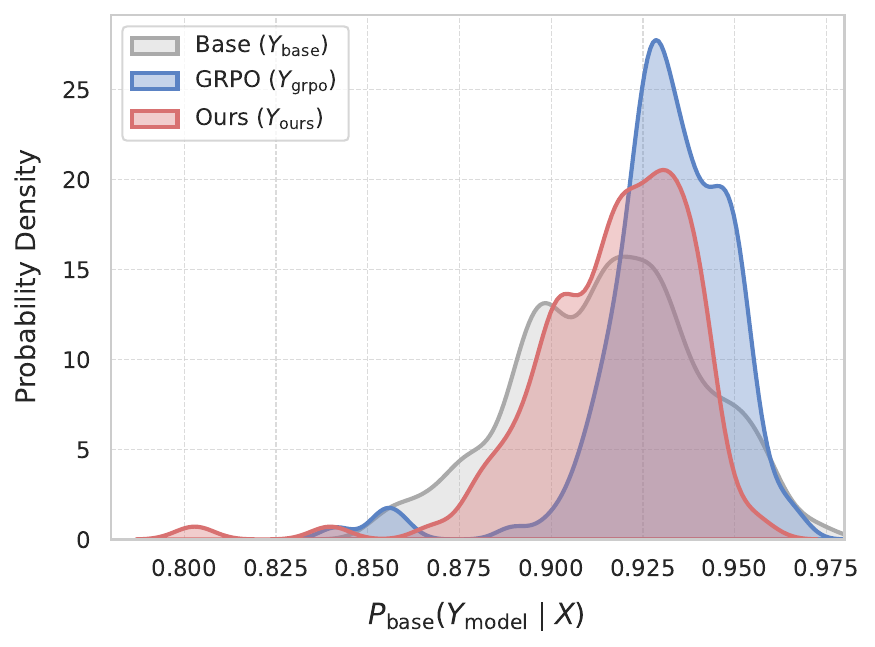}
    \caption{Kernel density estimation (KDE) of base-model probabilities for correct responses on AIME 2025 using Qwen2.5-Math-1.5B. For each correct response ($Y_{\text{model}}$) sampled from a policy model, we evaluate $P_{\text{base}}(Y_{\text{model}} \mid X)$. GRPO shifts the distribution toward high-probability responses under the base model,
    suggesting concentration on responses the base model already favors. Our method mitigates this concentration and preserves more low-probability correct responses.}
    \label{fig:base_likelihood_kde}
\end{figure}

To examine this change, Figure~\ref{fig:base_likelihood_kde} compares where correct responses from each policy lie under the base model's response distribution. For each correct response ($Y_{\text{model}}$) sampled from each policy model, we evaluate its probability under the base model and visualize these probabilities using kernel density estimation (KDE). The GRPO distribution shifts toward the high-probability region of the base model, indicating that GRPO concentrates on correct responses that the base model already generates with high probability. This suggests that the Pass@$k$ pattern is linked to concentration in the response distribution. GRPO improves sampling efficiency by increasing access to correct responses that are already likely under the base model, but reduces coverage of less likely correct responses. This motivates a closer look at how the GRPO update produces this concentration.


To understand how GRPO leads to this concentration, we examine two main components of its update, (i) which responses are sampled for each rollout group and (ii) how token-level gradients are computed. At the response level, we analyze which responses appear in the rollout group. Responses are sampled from the policy itself, making high-probability responses occur more often within a group. Because GRPO averages contributions across sampled responses equally, frequent responses accumulate larger total weight through repeated occurrence, causing the gradient mass to concentrate on them. At the token level, the GRPO importance ratio scales the gradient contribution of each sampled token. As the sampled token becomes more likely under the current policy, its update scale increases, further reinforcing the same local choice at a decision point. Over training, this feedback can make one choice increasingly dominant, reducing the alternative paths available to future rollouts. These two mechanisms make already likely reasoning paths increasingly dominant, reducing diversity.

To mitigate this issue, we propose \textit{Reweighting GRPO Against Distributional Concentration} (\textbf{ReCo}), with corresponding response-level and token-level corrections. At the response level, we normalize each response's gradient contribution by its expected count within the group, preventing frequent responses from dominating the group-level gradient. At the token level, we replace the standard token-level importance ratio with a variance-based ratio that accounts for both the sampled token probability and the probability mass left for alternatives. This downweights saturated choices and gives larger update scale to non-saturated decision points where multiple token choices remain plausible. Together, these corrections reduce GRPO's tendency to collapse onto frequent reasoning paths and help preserve reasoning diversity.
 
\paragraph{Contributions.} Our contributions are summarized as follows.
\begin{itemize}
    \item We analyze how GRPO concentrates updates on high-probability reasoning paths under the base model through response-level and token-level mechanisms.
    \item We introduce \textbf{ReCo}, a two-level correction for GRPO training. ReCo normalizes response-level updates by expected occurrence and replaces the token-level importance ratio with a variance-based ratio that places more update on positions where multiple token choices remain plausible, rather than on tokens that are already dominant.
    \item We evaluate ReCo on diverse mathematical reasoning benchmarks across 
    different model families and scales. ReCo consistently outperforms GRPO in 
    Pass@$k$ for large values of $k$ and response diversity while maintaining competitive performance for small values of $k$.
\end{itemize}

%% file: Sec/Preliminaries.tex
\section{Preliminaries}
\label{sec:preliminaries}

\subsection{RLVR Setup}

We consider Reinforcement Learning with Verifiable Rewards (RLVR) for post-training a language model policy $\pi_\theta$.
Given a prompt $q \sim \mathcal{D}$, the policy autoregressively generates a response
\[
    o = (o_1,\ldots,o_{|o|})
    \sim
    \pi_\theta(\cdot \mid q).
\]
RLVR assigns an outcome-based verifiable reward $R(q,o)$ to each completed response, computed by an external verifier.
Thus, rewards are defined at the sequence level rather than at the token level.



\subsection{Group Relative Policy Optimization}

For each prompt $q$, GRPO samples a group of $G$ responses from the old policy $\pi_{\theta_{\mathrm{old}}}$, the policy used to collect rollouts before the current update,
\[
    o_i \overset{\mathrm{i.i.d.}}{\sim} \pi_{\theta_{\mathrm{old}}}(\cdot \mid q),
    \quad i=1,\ldots,G.
\]
As in the RLVR setup, each response $o_i$ is an autoregressive sequence.
Each response receives a verifiable reward $R(q,o_i)$.
GRPO constructs a group-relative advantage
\[
    \hat{A}_i
    =
    \frac{R(q,o_i)-\mu_R}{\sigma_R},
\]
where
\[
    \mu_R = \frac{1}{G}\sum_{j=1}^{G} R(q,o_j), 
    \sigma_R =
    \mathrm{std}\big(\{R(q,o_j)\}_{j=1}^{G}\big).
\]
The same sequence-level advantage $\hat{A}_i$ is assigned to all tokens in response $o_i$.

GRPO updates the policy using token-level importance ratios. For the $t$-th token of response $o_i$, let
\[
    r_{i,t}(\theta)
    =
    \frac{
        \pi_\theta(o_{i,t} \mid q,o_{i,<t})
    }{
        \pi_{\theta_{\mathrm{old}}}(o_{i,t} \mid q,o_{i,<t})
    } .
\]
Following PPO-style clipping, define the clipped token objective
\[
\begin{aligned}
    \ell_{i,t}^{\mathrm{clip}}(\theta)
    =
    \min \Big(
    & r_{i,t}(\theta)\hat{A}_i, \\
    & \mathrm{clip}(r_{i,t}(\theta),1-\varepsilon,1+\varepsilon)
      \hat{A}_i
    \Big).
\end{aligned}
\]

The GRPO objective is then
\[
    \mathcal{J}_{\mathrm{GRPO}}(\theta)
    =
    \mathbb{E}
    \left[
    \frac{1}{G}
    \sum_{i=1}^{G}
    \frac{1}{|o_i|}
    \sum_{t=1}^{|o_i|}
    \ell_{i,t}^{\mathrm{clip}}(\theta)
    \right],
\]
where the expectation is over $q \sim \mathcal{D}$ and
$\{o_i\}_{i=1}^{G} \sim \pi_{\theta_{\mathrm{old}}}(\cdot \mid q)$.
In practice, GRPO is often optimized with an additional KL penalty to a reference policy $\pi_{\mathrm{ref}}$. To simplify the analysis, we omit clipping and auxiliary KL terms.
The resulting token-level policy-gradient contribution is
\[
    g_{i,t}^{\mathrm{GRPO}}
    =
    r_{i,t}(\theta)
    \nabla_\theta
    \log \pi_\theta(o_{i,t}\mid q,o_{i,<t})
    \hat{A}_i .
\]
Equivalently, the unclipped GRPO gradient can be written as
\[
\begin{aligned}
    \nabla_\theta
    \mathcal{J}_{\mathrm{GRPO}}(\theta)
    =
    \mathbb{E}
    \Bigg[
    \frac{1}{G}
    \sum_{i=1}^{G}
    \frac{1}{|o_i|}
    \sum_{t=1}^{|o_i|}
    g_{i,t}^{\mathrm{GRPO}}
    \Bigg].
\end{aligned}
\]

%% file: Sec/Method.tex


\section{Method}
\label{sec:method}

\subsection{Overview}
We propose \textit{Reweighting GRPO Against Distributional Concentration} (\textbf{ReCo}), a reweighting method that applies corrections to the two main components of the GRPO update, (i) which responses are sampled for each rollout group, and (ii) how token-level gradients are computed.


At the response level, ReCo normalizes each response's gradient contribution by its expected count within the rollout group.
Since high-probability responses are sampled more often, this normalization prevents them from dominating the group-level gradient through repeated occurrence.

At the token level, ReCo replaces the standard token-level importance
ratio with a variance-based ratio. This ratio accounts not only for the sampled
token probability but also for the remaining probability assigned to
alternative tokens. As a result, ReCo reduces update scale at saturated decision
points and gives relatively larger weight to tokens where alternative token choices remain plausible.

\subsection{Response-Level Count Reweighting}
\label{sec:method:response}

In GRPO, responses are not sampled uniformly from the space of possible solutions but from the old policy itself. As a result, responses with high probability under the old policy are more likely to appear multiple times within the group. Because the GRPO update averages over sampled responses, this repeated occurrence gives such trajectories a larger aggregate contribution.

ReCo uses expected occurrence to downweight responses that are likely to appear repeatedly. For a fixed prompt $q$ and a trajectory $o$, we define its occurrence count over the group as
\begin{equation}
    N_q(o) = \sum_{i=1}^{G} \mathbf{1}[o_i = o],
\end{equation}
whose expected value under repeated sampling from the old policy is
\begin{equation}
    \mathbb{E}[N_q(o)] = G\,\pi_{\theta_{\text{old}}}(o \mid q).
\end{equation}
This expected count provides an analytical measure of how frequently a trajectory is expected to be represented under the old policy.

One possible approach would be to directly count duplicated or semantically equivalent responses and normalize their contributions by the observed count. However, this is difficult in open-ended LLM generation where exact duplicate trajectories are rare and defining equivalence requires an external criterion. ReCo avoids this explicit grouping problem by using expected counts rather than observed equivalence counts.

This gives the inverse expected-count weight
\begin{equation}
    \frac{1}{G\,\pi_{\theta_{\text{old}}}(o_i \mid q)}.
\end{equation}
Under this weighting, a trajectory that is expected to appear multiple times in the rollout group receives a smaller per-sample weight, so that repeated representation does not by itself enlarge its aggregate contribution.

In practice, raw sequence likelihood depends strongly on response length, since it is a product of token probabilities smaller than one. Using raw likelihood directly would assign larger weights to longer responses simply because they contain more tokens. To reduce this length-induced bias, ReCo uses the
length-normalized likelihood
\begin{equation}
    \bar{\pi}_{\theta_{\text{old}}}(o_i \mid q) =
    \left(\prod_{t=1}^{|o_i|} \pi_{\theta_{\text{old}}}(o_{i,t} \mid h_{i,t})\right)^{1/|o_i|}
\end{equation}
as a practical proxy for how frequently a trajectory is generated. The response-level weight is then defined as
\begin{equation}
    w^{\text{resp}}_i = \frac{1}{G\,\bar{\pi}_{\theta_{\text{old}}}(o_i \mid q)}.
\end{equation}
The resulting response-level weight $w^{\text{resp}}_i$ reduces the per-sample
contribution of high-likelihood trajectories, mitigating the effect of frequent responses dominating the group gradient through repeated occurrence alone.

\subsection{Token-Level Variance Reweighting}

At the token level, GRPO's importance ratio creates a feedback loop that can amplify frequently sampled tokens. Specifically, GRPO assigns a larger gradient scale to sampled tokens whose probability has increased. From Section~\ref{sec:preliminaries}, the GRPO
token-level gradient contribution is
\[
    g_{i,t}^{\mathrm{GRPO}}
    =
    \frac{p_{i,t}^{\theta}}{p_{i,t}^{\text{old}}}
    \nabla_\theta
    \log \pi_\theta(o_{i,t}\mid h_{i,t})
    \hat{A}_i ,
\]
where
\[
    p_{i,t}^{\theta}
    =
    \pi_\theta(o_{i,t}\mid h_{i,t}),
    \quad
    p_{i,t}^{\text{old}}
    =
    \pi_{\theta_{\text{old}}}(o_{i,t}\mid h_{i,t}) .
\]
The ratio \(p_{i,t}^{\theta}/p_{i,t}^{\text{old}}\) grows when the sampled
token becomes more likely under the current policy. Thus, GRPO can repeatedly
amplify tokens that have been frequently selected at the same decision point,
making those choices even more likely in future rollouts. Over training, this
feedback can make a local choice increasingly dominant, reducing the alternative
paths available to future rollouts.

For reasoning diversity, an update should account for both the sampled token probability ($p$) and the probability mass left for alternatives ($1-p$). If a high-probability token is strongly reinforced, future rollouts are more likely to follow the same path, while alternative paths shrink. If a very low-probability token is updated, the absolute probability is still small, so the update has limited effect on future path diversity. In contrast, when the choice is not yet saturated, updates can more directly affect path diversity. Positive updates can strengthen a successful choice while alternatives remain available and negative updates can make future rollouts move more easily toward alternatives.

ReCo captures this with a variance ratio:
\[
    r_{i,t}^{\mathrm{var}}(\theta)
    =
    \frac{
        p_{i,t}^{\theta}(1-p_{i,t}^{\theta})
    }{
        p_{i,t}^{\text{old}}(1-p_{i,t}^{\text{old}})
    } .
\]




The term $p(1-p)$ corresponds to the Bernoulli variance of sampling a token with probability $p$. Here, \(p\) reflects whether the sampled token is likely to affect future rollouts, while \(1-p\) reflects how much probability mass remains for alternative choices. Their product is small when the sampled token is either too unlikely or already dominant, and large when the decision point remains non-saturated. Thus, the variance ratio assigns a larger scale to choices that can still influence reasoning-path diversity and avoids further amplifying saturated local decisions.

\subsection{Final Objective}
\label{sec:method:final}

Combining the response-level and token-level corrections, the ReCo token-level
gradient contribution is
\[
    g_{i,t}^{\mathrm{ReCo}}
    =
    w_i^{\mathrm{resp}}\,
    r_{i,t}^{\mathrm{var}}(\theta)\,
    \nabla_\theta
    \log \pi_\theta(o_{i,t}\mid h_{i,t})\,
    \hat{A}_i .
\]
The corresponding policy gradient is
\[
    \nabla_\theta
    \mathcal{J}_{\mathrm{ReCo}}(\theta)
    =
    \mathbb{E}
    \left[
    \frac{1}{G}
    \sum_{i=1}^{G}
    \frac{1}{|o_i|}
    \sum_{t=1}^{|o_i|}
    g_{i,t}^{\mathrm{ReCo}}
    \right] .
\]
Compared with standard GRPO, ReCo modifies the allocation of update magnitude at
both levels. The response-level weight $w_i^{\mathrm{resp}}$ reduces the aggregate
contribution from trajectories that are expected to appear frequently in the
rollout group, while the token-level variance ratio $r_{i,t}^{\mathrm{var}}(\theta)$
gives smaller scale to saturated decisions and larger scale to non-saturated
decision points where alternative token choices remain plausible.

\input{Sec/tables/table1}

For implementation with PPO-style clipping, we replace $r_{i,t}^{\mathrm{var}}(\theta)$
with
\[
    \tilde{r}_{i,t}^{\mathrm{var}}(\theta)
    =
    \frac{
        p_{i,t}^{\theta}\,
        \big(1-\mathrm{sg}(p_{i,t}^{\theta})\big)
    }{
        p_{i,t}^{\text{old}}\,
        \big(1-p_{i,t}^{\text{old}}\big)
    },
\]

where $\mathrm{sg}(\cdot)$ denotes stop-gradient. Applying stop-gradient to the
$(1-p_{i,t}^{\theta})$ factor ensures that the variance ratio acts as a reweighting scalar under clipping, keeping the clipping mechanism aligned with standard PPO behavior. The clipped ReCo token objective is
\begin{align*}
    \ell_{i,t}^{\mathrm{ReCo}}(\theta)
    =\;& w_i^{\mathrm{resp}}
    \min \Big(
       \tilde{r}_{i,t}^{\mathrm{var}}(\theta)\,\hat{A}_i, \\
    & \quad\quad\;
    \mathrm{clip}\big(
        \tilde{r}_{i,t}^{\mathrm{var}}(\theta),\,
        1-\varepsilon,\, 1+\varepsilon
    \big)\hat{A}_i
    \Big).
\end{align*}
The final ReCo objective is
\[
    \mathcal{J}_{\mathrm{ReCo}}(\theta)
    =
    \mathbb{E}
    \left[
    \frac{1}{G}
    \sum_{i=1}^{G}
    \frac{1}{|o_i|}
    \sum_{t=1}^{|o_i|}
    \ell_{i,t}^{\mathrm{ReCo}}(\theta)
    \right] .
\]
We keep the remaining training components, such as the KL penalty to a
reference policy $\pi_{\mathrm{ref}}$, the same as in the GRPO baseline.

%% file: Sec/tables/table1.tex
\begin{table*}[t]
    \centering
    \small
    \setlength{\tabcolsep}{12pt}
    \renewcommand{\arraystretch}{1}
    
    \definecolor{modelbg}{RGB}{245, 246, 248}
    \definecolor{recobg}{RGB}{232, 242, 253}
    \definecolor{deltapos}{RGB}{25, 95, 180}
    
    \newcommand{\modelheader}[1]{%
        \rowcolor{modelbg}
        \multicolumn{7}{>{\centering\arraybackslash}c}{\textbf{#1}} \\
        \addlinespace[2pt]
    }
    
    \caption{
        Evaluation results on five mathematical reasoning benchmarks (Pass@64, \%).
        \textbf{Bold} = best, \underline{underline} = second best.
    }
    \label{tab:main_math_results_pass64}
    \begin{tabular}{lcccccc}
        \toprule
        \textbf{Method}
            & \textbf{AIME '25}
            & \textbf{AIME '24}
            & \textbf{AMC '23}
            & \textbf{Olympiad}
            & \textbf{MATH-500}
            & \textbf{Avg.} \\
        \midrule\midrule
        
        \modelheader{Qwen2.5-Math-1.5B}
        Base                  & 32.0 & 39.1 & 89.3 & \underline{64.0} & \underline{92.9} & 63.5 \\
        GRPO                  & \underline{36.2} & \underline{46.5} & \underline{91.3} & 63.1 & 92.0 & \underline{65.8} \\
        \rowcolor{recobg}
        \textbf{ReCo (Ours)}  & \textbf{40.1} & \textbf{48.9} & \textbf{93.4} & \textbf{68.0} & \textbf{94.1} & \textbf{68.9} \\
        \textit{$\Delta$ over GRPO}
            & \textcolor{deltapos}{\textit{+3.9}}
            & \textcolor{deltapos}{\textit{+2.4}}
            & \textcolor{deltapos}{\textit{+2.1}}
            & \textcolor{deltapos}{\textit{+4.9}}
            & \textcolor{deltapos}{\textit{+2.1}}
            & \textcolor{deltapos}{\textit{+3.1}} \\
        \addlinespace[4pt]
        \midrule
        
        \modelheader{Qwen2.5-Math-7B}
        Base                  & 34.7 & 55.7 & 89.8 & \underline{67.7} & \underline{94.1} & 68.4 \\
        GRPO                  & \underline{39.9} & \underline{59.4} & \underline{87.9} & 65.0 & 92.8 & \underline{69.0} \\
        \rowcolor{recobg}
        \textbf{ReCo (Ours)}  & \textbf{42.7} & \textbf{62.8} & \textbf{94.3} & \textbf{68.4} & \textbf{94.7} & \textbf{72.6} \\
        \textit{$\Delta$ over GRPO}
            & \textcolor{deltapos}{\textit{+2.8}}
            & \textcolor{deltapos}{\textit{+3.4}}
            & \textcolor{deltapos}{\textit{+6.4}}
            & \textcolor{deltapos}{\textit{+3.4}}
            & \textcolor{deltapos}{\textit{+1.9}}
            & \textcolor{deltapos}{\textit{+3.6}} \\
        \addlinespace[4pt]
        \midrule
        
        \modelheader{Llama-3.1-8B-Instruct}
        Base                  & \underline{21.5} & \underline{36.5} & \textbf{90.8} & \textbf{60.1} & \textbf{89.5} & \textbf{59.7} \\
        GRPO                  & \phantom{0}6.7 & 36.2 & 66.3 & 44.4 & 77.3 & 46.2 \\
        \rowcolor{recobg}
        \textbf{ReCo (Ours)}  & \textbf{21.9} & \textbf{41.0} & \underline{81.0} & \underline{55.6} & \underline{88.1} & \underline{57.5} \\
        \textit{$\Delta$ over GRPO}
            & \textcolor{deltapos}{\textit{+15.2}}
            & \textcolor{deltapos}{\textit{+4.8}}
            & \textcolor{deltapos}{\textit{+14.7}}
            & \textcolor{deltapos}{\textit{+11.2}}
            & \textcolor{deltapos}{\textit{+10.8}}
            & \textcolor{deltapos}{\textit{+11.3}} \\
        \bottomrule
    \end{tabular}
\end{table*}

%% file: Sec/Experiments.tex
\begin{figure*}[t]
    \centering
    \includegraphics[width=\textwidth]{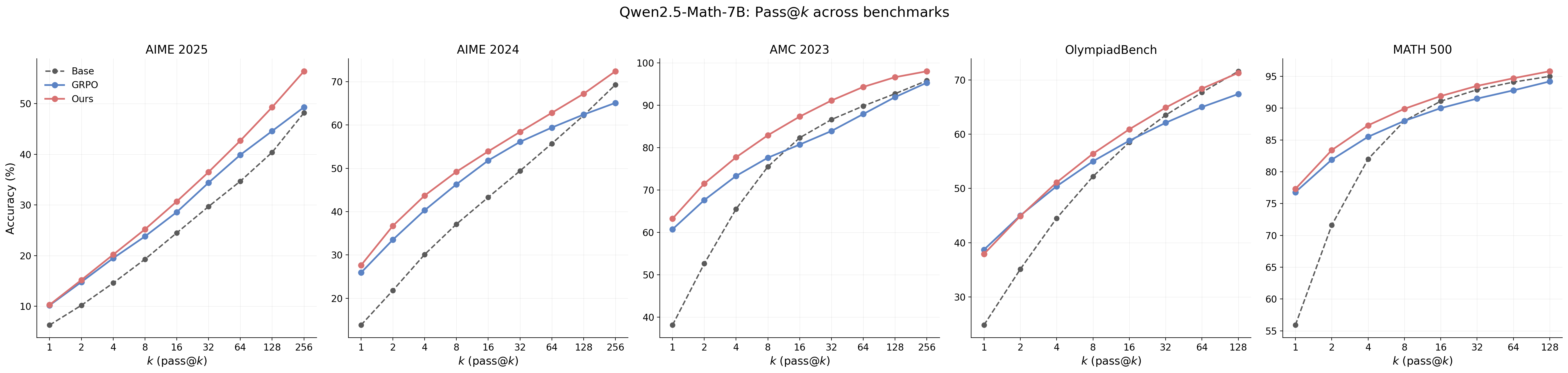}
    \caption{
    Pass@$k$ scaling curves on five mathematical reasoning benchmarks 
    for Qwen2.5-Math-7B. Each panel shows Base, GRPO, and ReCo (Ours) 
    across different sampling budgets $k$.
    }
    \label{fig:passk_curves}
\end{figure*}

\section{Experiments}
\label{sec:experiments}

\subsection{Experimental Setup}
\label{sec:experimental_setup}

\paragraph{Training Setup.}

We train all RL methods on the MATH dataset~\citep{hendrycks2021MATH} using the
verl framework~\citep{Sheng_2025_VERL}. To compare algorithms across different model
scales and families, we conduct experiments with three backbone models, Qwen2.5-Math-1.5B/7B~\citep{qwen_model}, and Llama-3.1-8B-Instruct~\citep{llama_model}. We use a prompt batch size of $256$, generate 8 rollouts per prompt with temperature $1.0$ and top-$p$ $1.0$, use a PPO mini-batch size of $64$, and set the learning rate to $1\times 10^{-6}$. We compare ReCo against GRPO and keep all remaining hyperparameters identical for a fair comparison. Additional hyperparameters and implementation details are provided in
Appendix~\ref{sec:appendix_training_details}.

\paragraph{Evaluation Setup.}

We evaluate on five mathematical reasoning benchmarks, AIME 2025, AIME 2024, AMC 2023, OlympiadBench~\citep{he2024olympiadbench}, and
MATH 500~\citep{hendrycks2021MATH}. Following prior work~\citep{yue2025does}, we sample responses with temperature $0.6$ and top-$p$
$0.95$ during evaluation.

Our primary metric is Pass@$k$ accuracy~\citep{chen2021evaluatinglargelanguagemodels}. Lower values of $k$ measure whether training improves sampling efficiency over the base model, while higher values of $k$ measure whether the trained policy retains, loses, or expands the base model's ability to produce correct solutions. Further details are provided in Appendix~\ref{sec:eval_details}.


\subsection{Main Results}
\label{sec:main_results}

\paragraph{Performance across Models and Benchmarks.}

Table~\ref{tab:main_math_results_pass64} reports Pass@64 across five benchmarks
and three backbone models. We use Pass@64 as a proxy for reasoning capacity at large sampling budgets. ReCo improves over GRPO on every benchmark and backbone. On both Qwen2.5-Math-1.5B and Qwen2.5-Math-7B, ReCo gives consistent gains across all five benchmarks, improving the average score by $+3.1$ and $+3.6$ points, respectively. The largest improvement is $+6.4$ points on AMC 2023 with Qwen2.5-Math-7B. Llama-3.1-8B-Instruct results show a different pattern. Consistent with prior reports~\citep{yue2025does, zhu202nsr, OctoThinker, yeo2025demystifying} that RL training on this backbone can underperform the base model under Pass@$k$ evaluation, GRPO falls below the base model on all five benchmarks. ReCo reduces this degradation, improving over GRPO on every benchmark by $+11.3$ points on average and even surpassing the base model on AIME 2025 and AIME 2024.



\paragraph{Pass@$k$ across Different Values of $k$.}

Figure~\ref{fig:passk_curves} evaluates Pass@$k$ across different values of $k$
on Qwen2.5-Math-7B. At small $k$, ReCo is comparable to GRPO, suggesting that it preserves the sampling-efficiency gains of RL training. At larger $k$, ReCo consistently outperforms GRPO across every benchmark. In several cases, ReCo also exceeds the base model at large $k$, indicating that it 
achieves broader coverage of correct reasoning trajectories than 
the base model itself.



\begin{figure*}[t]
    \centering
    \includegraphics[width=0.95\linewidth]{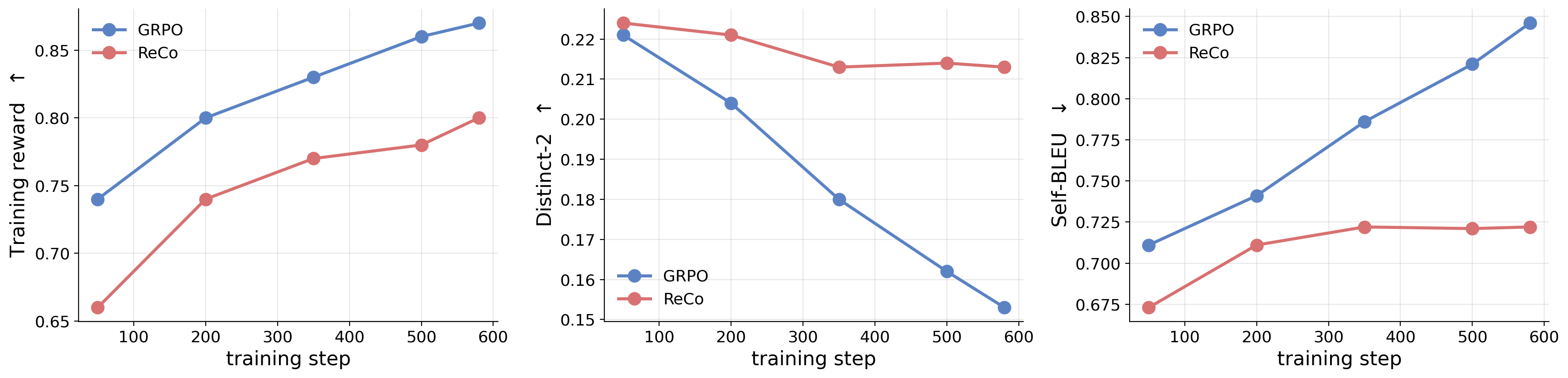}
    \caption{
    Training dynamics of training reward, Distinct-2, and Self-BLEU 
    over training steps on Qwen2.5-Math-1.5B. Higher Distinct-2 and lower 
    Self-BLEU indicate greater response diversity.
    }
    \label{fig:reward_diversity}
\end{figure*}

\subsection{Further Analysis}
\label{sec:further_analysis}

\paragraph{Training Dynamics of Reward and Rollout Diversity.}

We analyze how the rollout distribution changes during training. Using
intermediate checkpoints from RL training initialized with Qwen2.5-Math-1.5B,
we generate rollouts on the MATH training set and measure their average reward, Distinct-2~\citep{li2016diversity}, and Self-BLEU~\citep{papineni-etal-2002-bleu}. This analysis examines whether the policy continues to produce diverse rollouts on the training prompts as reward
improves.

Figure~\ref{fig:reward_diversity} shows that GRPO increases the training reward rapidly while losing diversity. Distinct-2 consistently decreases, while Self-BLEU consistently increases, indicating that GRPO rollouts become increasingly similar over
training. This suggests that GRPO concentrates probability mass on a narrower set of high-reward responses, which improves the reward but reduces the variety of reasoning paths explored during training.


ReCo shows a different trend. Although its reward improves more gradually, it keeps Distinct-2 and Self-BLEU much closer to their initial values. The slower reward growth does not reflect a deficit in learning, since Table~\ref{tab:main_math_results_pass64} shows that ReCo achieves higher Pass@$k$ than GRPO across all backbones and benchmarks at the end of training. This broader exploration during training is consistent with the improved large-$k$ behavior observed in Section~\ref{sec:main_results}.

\begin{figure}[t]
    \centering
    \includegraphics[width=\linewidth]{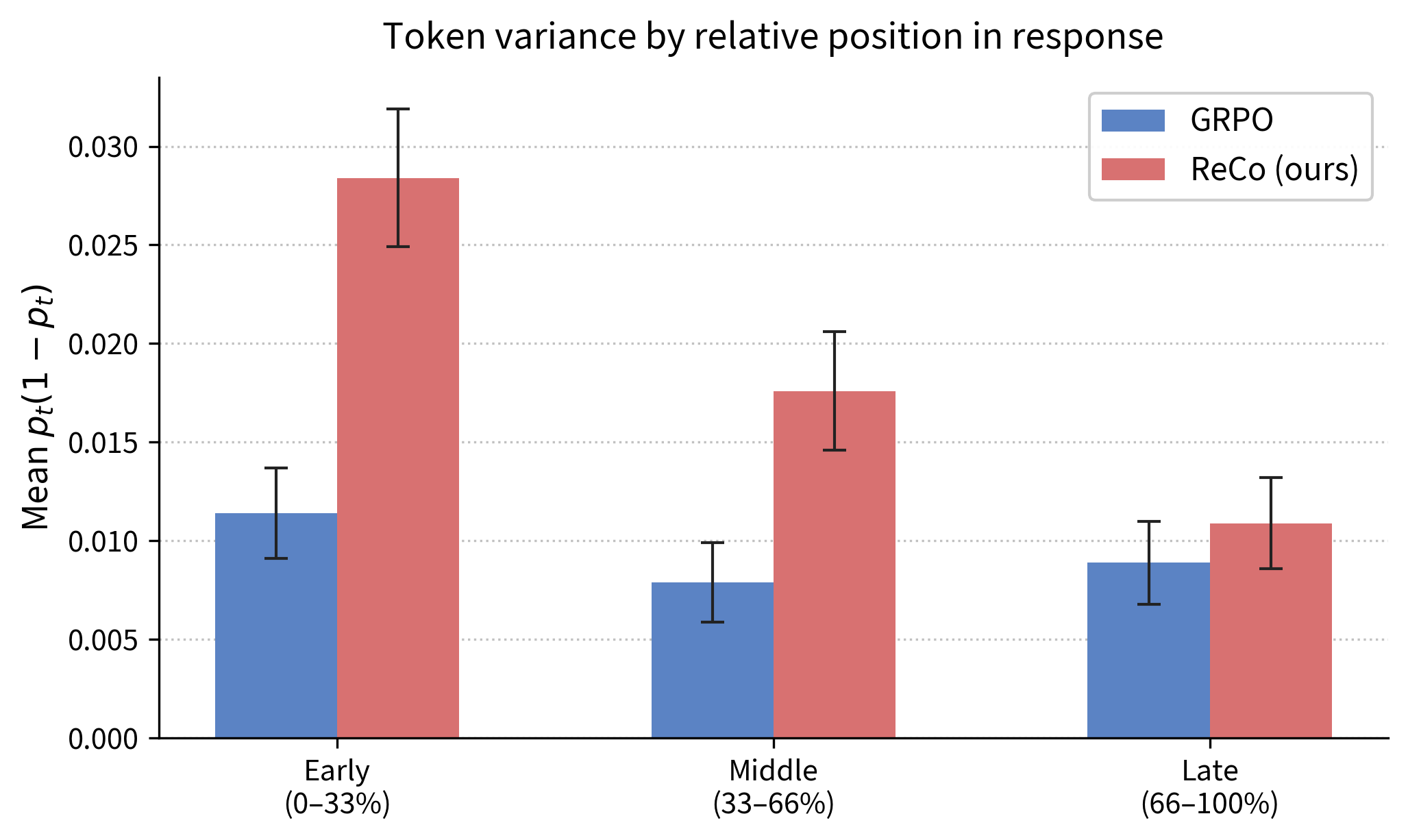}
    \caption{
    Mean token-level Bernoulli variance $p_t(1-p_t)$ by relative position 
    within each response on Qwen2.5-Math-1.5B. Error bars show standard 
    error across responses. Higher values indicate that the sampled token 
    has more room for alternative choices.
    }
    \label{fig:token_variance}
\end{figure}
\paragraph{Token Variance across Response Positions.}
To examine ReCo's token-level effect on diversity, we measure the Bernoulli variance $p_t(1-p_t)$ of the sampled token probability 
across positions within each response. Since $p_t(1-p_t)$ is large when the token choice is non-saturated (i.e., $p$ is away from 0 
or 1) and is the quantity ReCo reweights by 
(Section~\ref{sec:method}), higher values indicate that more positions remain non-saturated under the trained policy. We group 
token positions into early ($0$--$33\%$), middle ($33$--$66\%$), and late ($66$--$100\%$) segments relative to the full response length. Figure~\ref{fig:token_variance} shows the mean $p_t(1-p_t)$ for each segment under GRPO and ReCo. ReCo maintains substantially higher token variance than GRPO in the early 
and middle segments (2.5$\times$ and 2.2$\times$ higher, respectively), 
while the two methods converge in the late segment. The early and middle segments, where reasoning paths typically branch into different strategies, retain meaningful branching under ReCo, while GRPO drives them toward more deterministic choices. The late segment, which corresponds to answer-confirmation steps, naturally saturates under both methods.

\input{Sec/tables/table_diversity_eval}
\input{Sec/tables/compare_token_response}

\paragraph{Inference-time Diversity.}
We further compare the diversity of correct responses generated at inference
time. Using Qwen2.5-Math-1.5B, we sample 16 responses per problem on AIME 2025, AIME 2024, and AMC 2023, and compute the diversity among correct responses using $1-\mathrm{Self\text{-}BLEU}$, which captures surface-level lexical variation. Higher values indicate greater
diversity. As shown in Table~\ref{tab:diversity_eval}, ReCo achieves higher diversity than GRPO on all three benchmarks, indicating that ReCo produces less repetitive reasoning traces among correct generations. These results complement the Pass@$k$ analysis by showing that ReCo's improved coverage also extends to inference time.

\paragraph{Effect of Each Correction.}
We ablate the components of ReCo in Table~\ref{tab:ablation_components}. \textbf{Response only} applies expected-occurrence normalization at the response level, and \textbf{Token only} applies the variance-based token-level ratio.

\textit{The two corrections are complementary.} Both single-correction variants improve over GRPO on average, and their combination achieves the best average performance. Response-level reweighting changes how sampled responses are aggregated within the rollout group, while token-level reweighting changes how update scale is distributed across token decisions inside a response. The two thus act on different aspects of the update, consistent with the gains from combining them.

\textit{Variance-based token reweighting works best.} We compare token-level reweighting forms. GRPO uses $p_{\theta}/p_{\text{old}}$, which gives larger scale to tokens whose probabilities have increased. ASPO~\citep{wang2026aspo} uses $p_{\text{old}}/p_{\theta}$, which gives larger scale to low-probability sampled tokens. ReCo instead uses $p_{\theta}(1-p_{\theta})/p_{\text{old}}(1-p_{\text{old}})$, which emphasizes non-saturated choices by considering both the sampled token probability and the remaining probability mass for alternatives. Its stronger performance provides empirical support for $p_{\theta}(1-p_{\theta})$ as the token-level reweighting criterion.

%% file: Sec/tables/table_diversity_eval.tex
\begin{table}[htbp]
    \centering
    \setlength{\tabcolsep}{4pt}
    \renewcommand{\arraystretch}{1.2}
    
    \definecolor{recobg}{RGB}{232, 242, 253} 
    
    \caption{
        Dataset-level diversity evaluation ($1 - \text{Self-BLEU}$) across mathematical reasoning benchmarks.
        Each value is computed at the dataset level. Higher values indicate greater diversity.
    }
    \label{tab:diversity_eval}
    
    \resizebox{\columnwidth}{!}{%
    \begin{tabular}{lccc}
        \toprule
        \textbf{Setting} 
            & \textbf{AIME 2025} 
            & \textbf{AIME 2024} 
            & \textbf{AMC 2023} \\
        \midrule\midrule
        GRPO                  & 0.2590 & 0.3513 & 0.2148 \\
        \rowcolor{recobg}
        \textbf{ReCo (Ours)}  & \textbf{0.2686} & \textbf{0.4184} & \textbf{0.2602} \\
        \bottomrule
    \end{tabular}%
    }
\end{table}

%% file: Sec/tables/compare_token_response.tex
\begin{table*}[t]
    \centering
    \small
    \setlength{\tabcolsep}{9pt} 
    \renewcommand{\arraystretch}{1.1} 
    
    \definecolor{recobg}{RGB}{232, 242, 253}   
    
    \caption{
        Ablation on the two corrections of ReCo (Pass@64 on Qwen2.5-Math-1.5B).
        \textbf{Response} applies only the response-level expected-occurrence 
        normalization. \textbf{Token} applies only the token-level variance 
        ratio. \textbf{ReCo} combines both. \textbf{Bold} indicates the best 
        result.
    }
    \label{tab:ablation_components}
    \begin{tabular}{lcccccc}
        \toprule
        \textbf{Method}
            & \textbf{AIME 25}
            & \textbf{AIME 24}
            & \textbf{AMC 23}
            & \textbf{Olympiad}
            & \textbf{MATH 500}
            & \textbf{Avg.} \\
        \midrule\midrule
        GRPO              & 36.2 & 46.5& 91.3 & 63.1 & 92.0 & 65.8 \\
        ASPO              & 37.9 & 47.6 & 89.1 & 63.2 & 92.0 & 66.0 \\
        \midrule
        Response only     & \underline{38.9} & \underline{49.2} & \textbf{93.7} & 65.4 & \underline{93.5} & \underline{68.1} \\
        Token only        & 38.3 & \textbf{50.5} & 91.2 & \underline{65.8} & 92.5 & 67.7 \\
        \rowcolor{recobg}
        \textbf{ReCo}     & \textbf{40.1} & 48.9 & \underline{93.4} & \textbf{68.0} & \textbf{94.1} & \textbf{68.9} \\
        \bottomrule
    \end{tabular}
\end{table*}

%% file: Sec/Related_work.tex
\section{Related Work}
\label{sec:related_work}

\paragraph{RLVR and Reasoning Capacity.}
RLVR has become a common post-training paradigm for enhancing LLM reasoning. Recent reasoning models and open RL systems have shown that outcome-based RL can substantially improve reasoning performance~\citep{jaech2024openai,guo2025deepseekr1,
yu2025dapo,zhang2025srpo}. GRPO is a representative RLVR method that optimizes a policy using groups of sampled responses and outcome-level rewards, avoiding a 
separate value model~\citep{shao2024deepseekmath}. Building on this, subsequent work has studied R1-zero-like training~\citep{liu2025understanding}. Recent analyses question whether RLVR truly expands the base model's reasoning capacity or mainly improves sampling efficiency over already accessible paths.~\citet{yue2025does,wu2026invisibleleashrlvrescape} show that RLVR-trained models can improve at small sampling budgets but fail to surpass the base model at large Pass@$k$, suggesting limited exploration beyond existing reasoning paths. Our work studies this phenomenon from the perspective of the GRPO update itself, identifying response-level and token-level mechanisms that drive distributional concentration.

\paragraph{Diversity in RLVR.}
A related line of work studies how reasoning-oriented fine-tuning changes the coverage of generated solution paths. Prior work observes that gains in single-sample accuracy can coincide with reduced coverage under repeated sampling~\citep{dang2024diversitycollapse,yue2025does,wu2026invisibleleashrlvrescape}. Recent methods address this issue by encouraging diversity ~\citep{li2025jointlyreinforcingdiversityquality,
yao2025diversityawarepolicyoptimizationlarge,
chen2025dragrpo}, or by studying alternative divergence choices~\citep{li2025divergence}. ReCo differs in approach. Rather than adding an external diversity objective or regularizer, it reweights the GRPO update itself based on how concentration arises within the update, where frequent trajectories receive larger response-level gradient mass through repeated sampling and token-level importance weighting amplifies saturated choices.

%% file: Sec/Conclusion.tex
\section{Conclusion}
\label{sec:conclusion}

We studied how GRPO changes the sampling behavior of reasoning models. We found that GRPO can concentrate probability mass on responses that are already frequent under the base model, which improves Pass@$k$ at lower values of $k$ but can reduce coverage at higher values of $k$. We identify two mechanisms behind this effect. At the response level, frequent responses receive larger aggregate gradient mass through repeated occurrence, while at the token level, importance ratios amplify choices whose probabilities have already increased.

To mitigate these mechanisms, we introduced \textbf{ReCo}, a two-level reweighting method.
It normalizes response-level updates by expected occurrence and replaces the token-level importance ratio with a variance-based ratio that downweights saturated decision points while giving larger scale to non-saturated ones where multiple token choices remain plausible.

Empirically, ReCo improves Pass@$k$ at higher values of $k$ over GRPO while maintaining competitive performance at lower values of $k$.
It also preserves rollout diversity during training and generates more diverse correct responses at inference time. These results indicate that addressing GRPO's internal concentration mechanisms preserves a broader set of correct reasoning paths without sacrificing sampling efficiency.

%% file: Sec/Appendix.tex
\label{sec:appendix}

\section{Implementation Details}
\label{sec:implementation_details}

\subsection{Training Details}
\label{sec:appendix_training_details}

We train all RL methods on the MATH training set~\citep{hendrycks2021MATH} using the verl framework~\citep{Sheng_2025_VERL}. Table~\ref{tab:training_hyperparams} summarizes the training hyperparameters, which are kept identical for ReCo and GRPO. 

\begin{table}[h]
    \centering
    \small
    \setlength{\tabcolsep}{4pt}
    \renewcommand{\arraystretch}{1.1}
    \caption{Training hyperparameters.}
    \label{tab:training_hyperparams}
    \begin{tabular}{ll}
        \toprule
        \textbf{Hyperparameter} & \textbf{Value} \\
        \midrule
        Framework & verl \\
        Training data & MATH \\
        Optimizer & AdamW \\
        Learning rate & $1 \times 10^{-6}$ \\
        Prompt batch size & 256 \\
        PPO mini-batch size & 64 \\
        Rollouts per prompt ($G$) & 8 \\
        PPO update epochs & 1 \\
        PPO clip range ($\varepsilon$) & 0.2 \\
        Response-weight clip (ReCo only) & 10 \\
        KL coefficient ($\beta$) & 0.001 \\
        Sampling temperature & 1.0 \\
        Top-$p$ & 1.0 \\
        Reward & Binary correctness \\
        \midrule
        Total epochs & \\
        \quad Qwen 1.5B, Qwen 7B & 20 \\
        \quad Llama 8B & 8 \\
        \midrule
        Max response length & \\
        \quad Qwen 1.5B & 2048 \\
        \quad Qwen 7B, Llama 8B & 3072 \\
        \midrule
        GPUs & \\
        \quad Qwen 1.5B & $2\times$ A6000 \\
        \quad Qwen 7B, Llama 8B & $2\times$ B200 \\
        \bottomrule
    \end{tabular}
\end{table}

\paragraph{Response-Weight Clipping.}
$w_i^{\mathrm{resp}}$ can be very large when
$\bar{\pi}_{\theta_{\mathrm{old}}}(o_i \mid q)$ is small.
We clip $w_i^{\mathrm{resp}}$ to a maximum of 10 to prevent rare
extreme weights from producing unstable updates.
In practice, the cap is triggered in 10 out of 580 training steps (1.7\%), all during the early stage of training.
We additionally train ReCo without response-weight clipping while
keeping all other settings fixed.
As shown in Table~\ref{tab:response_clip}, removing the cap yields
performance close to the default setting, with an average Pass@64 of
68.6 compared with 68.9.
This indicates that ReCo's performance is not sensitive to the
clipping threshold and that the cap mainly serves as a safeguard
against rare extreme weights.

\begin{table}[t]
\centering
\small
\begin{tabular}{lcc}
\toprule
Benchmark
& ReCo, default
& No clipping \\
\midrule
AIME 25    & \textbf{40.1} & 39.5 \\
AIME 24    & \textbf{48.9} & \textbf{48.9} \\
AMC 23     & \textbf{93.4} & 93.2 \\
Olympiad    & \textbf{68.0} & 67.9 \\
MATH 500    & \textbf{94.1} & 93.5 \\
\midrule
Average     & \textbf{68.9} & 68.6 \\
\bottomrule
\end{tabular}
\caption{
Sensitivity to response-weight clipping on
Qwen2.5-Math-1.5B, measured by Pass@64 (\%).
Removing the response-weight cap produces performance
close to the default ReCo setting.
}
\label{tab:response_clip}
\end{table}

\subsection{Evaluation Details}
\label{sec:eval_details}
We use temperature $0.6$, top-$p$ $0.95$, and a maximum generation length of $16{,}384$ tokens, following prior work~\cite{yue2025does}. For Qwen-based models, we use
the prompt format shown in Figure~\ref{fig:qwen_prompt_template},
following~\cite{zeng2025simplerlzoo}. For Llama-based models, we apply the corresponding model
chat template.

\begin{figure}[t]
\centering
\begin{tcolorbox}[
    colback=gray!5,
    colframe=gray!35,
    boxrule=0.5pt,
    arc=2pt,
    width=0.92\linewidth,
    left=6pt,
    right=6pt,
    top=5pt,
    bottom=5pt
]
\small
\texttt{<|im\_start|>system}\\
\texttt{You are a helpful assistant.}\\
\texttt{<|im\_end|>}\\[1pt]
\texttt{<|im\_start|>user}\\
\texttt{\{input\}}\\
\texttt{Please reason step by step, and put your final answer within \textbackslash boxed\{\}.}\\
\texttt{<|im\_end|>}\\[1pt]
\texttt{<|im\_start|>assistant}
\end{tcolorbox}
\caption{Prompt format used for Qwen-based models.}
\label{fig:qwen_prompt_template}
\end{figure}

We report Pass@$k$ with $k \in \{1,2,4,8,\ldots,256\}$ for AIME 2024, AIME 2025, and AMC 2023, and $k \in \{1,2,4,8,\ldots,128\}$ for MATH500 and OlympiadBench. We estimate Pass@$k$ using the unbiased estimator from~\cite{chen2021evaluatinglargelanguagemodels}, with $n=1024$ samples per problem for AIME 2024, AIME 2025, and AMC 2023, and $n=128$ for MATH500 and OlympiadBench. Given $c$ correct samples among $n$ generations, the estimate is computed as

\begin{equation}
{\mathrm{Pass@}k}
=
\mathbb{E}_{x \sim \mathcal{D}}
\left[
1 -
\frac{\binom{n-c}{k}}{\binom{n}{k}}
\right],
\quad n \geq k.
\end{equation}
    
\section{Additional Experiments}
\label{sec:additional_experiments}
\subsection{Pass@k Across Models}

\begin{figure*}[t]
    \centering
    \includegraphics[width=\textwidth]{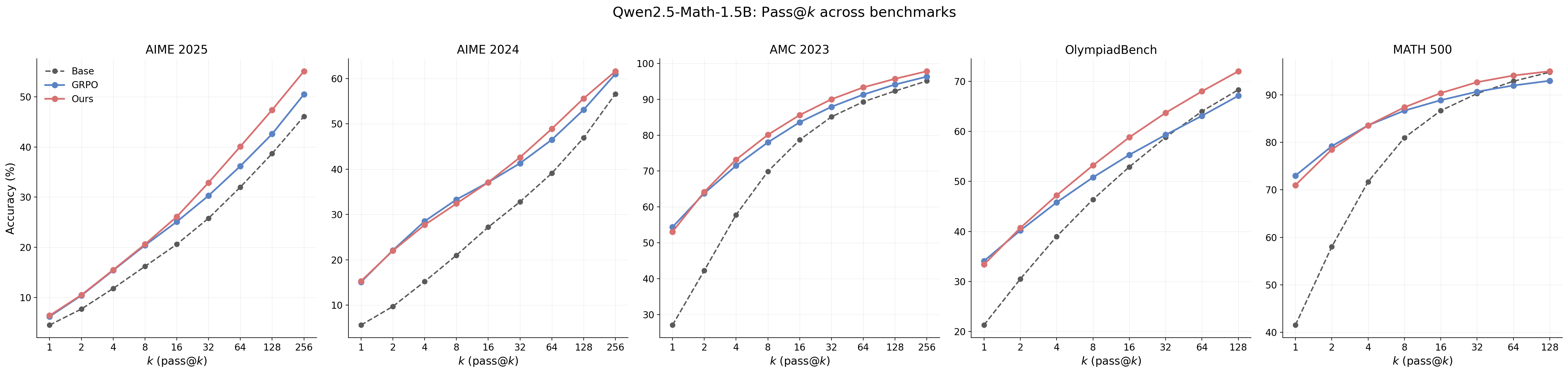}
    \caption{
    Pass@$k$ scaling curves on five mathematical reasoning benchmarks 
    for Qwen2.5-Math-1.5B. Each panel shows Base, GRPO, and ReCo (Ours) 
    across different sampling budgets $k$.
    }
    \label{fig:passk_curves_1.5b}
\end{figure*}

\begin{figure*}[t]
    \centering
    \includegraphics[width=\textwidth]{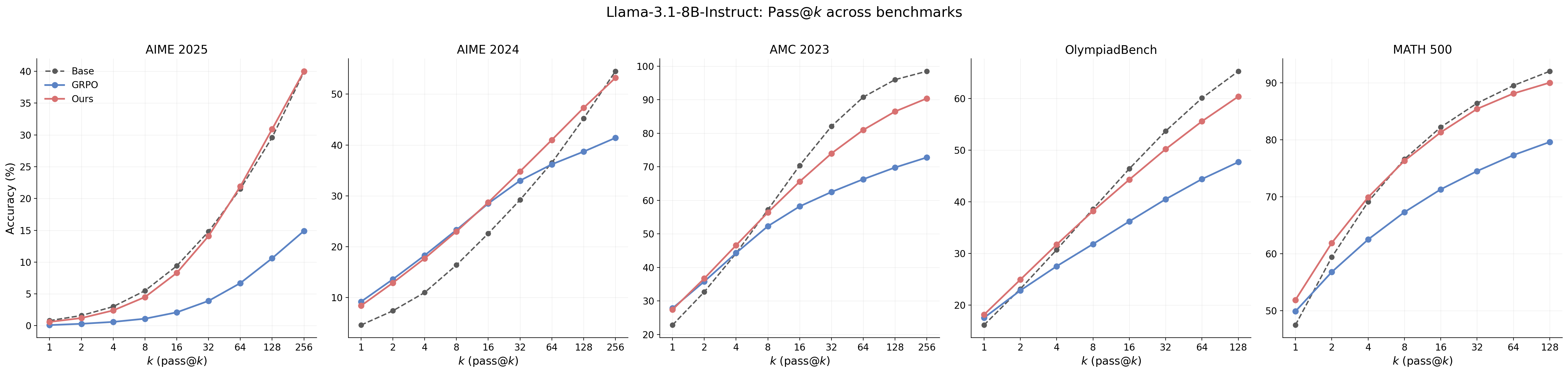}
    \caption{
    Pass@$k$ scaling curves on five mathematical reasoning benchmarks 
    for Llama-3.1-8B-Instruct. Each panel shows Base, GRPO, and ReCo (Ours) 
    across different sampling budgets $k$.
    }
    \label{fig:passk_curves_8b}
\end{figure*}

Figure~\ref{fig:passk_curves_1.5b} shows the Pass@$k$ scaling curves for Qwen2.5-Math-1.5B. ReCo and GRPO are comparable at small $k$, while ReCo outperforms GRPO as $k$ increases, indicating broader coverage of correct reasoning trajectories than GRPO.

Figure~\ref{fig:passk_curves_8b} shows the corresponding curves for Llama-3.1-8B-Instruct. Consistent with prior reports that RL training on this backbone can underperform the base model under Pass@$k$ evaluation~\citep{yue2025does, zhu202nsr, OctoThinker, yeo2025demystifying}, GRPO falls well below the base model across all benchmarks. ReCo substantially reduces this gap and closely tracks the base model on most benchmarks, while generally matching GRPO at small $k$ and outperforming it at larger sampling budgets.

\subsection{Reasoning Path Uniqueness}
\label{app:path_uniqueness}

We further measure how reasoning path diversity evolves during training. 
For each prompt at a given training step, we generate 16 rollouts and compute the ratio of unique reasoning prefixes within the group. 
Following recent findings that the early tokens of a reasoning trace 
carry disproportionately strong signals about the eventual solution 
trajectory~\citep{ji2026first}, we use the first $300$ 
characters of each rollout as the reasoning prefix. A uniqueness ratio 
of $1.0$ indicates that all 16 rollouts begin with distinct reasoning approaches, a lower ratio indicates that multiple rollouts share the same opening, suggesting that they will follow the same reasoning trajectory.

\begin{figure}[t]
    \centering
    \includegraphics[width=\linewidth]{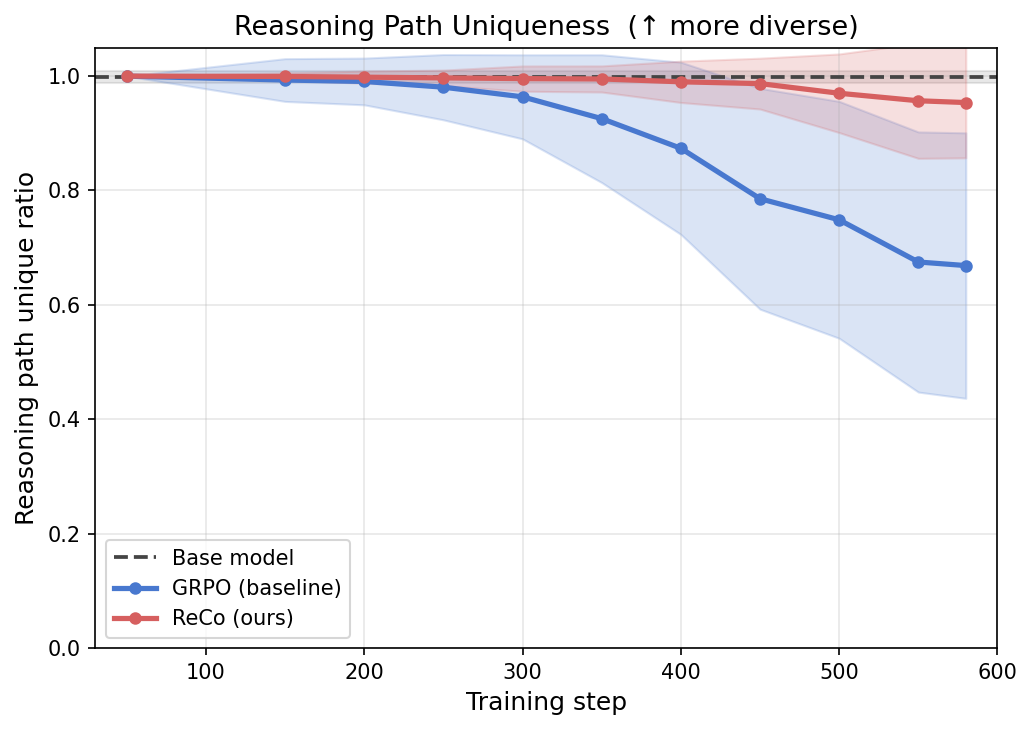}
    \caption{Reasoning path uniqueness across training steps. Higher 
    values indicate more diverse reasoning prefixes among rollouts. 
    GRPO progressively loses path diversity, while ReCo preserves it 
    close to the base model.}
    \label{fig:path_uniqueness}
\end{figure}

Figure~\ref{fig:path_uniqueness} shows this ratio across training steps 
for Qwen2.5-Math-7B. Both methods start near $1.0$, matching the base 
model's full diversity. As training progresses, GRPO's reasoning path 
uniqueness drops to $0.66$ by step 580, indicating that roughly one-third 
of rollouts begin with the same reasoning approach as another rollout 
in the group. ReCo maintains a uniqueness ratio of $0.95$, preserving 
nearly the full reasoning diversity of the base model.

\begin{table*}[t]
\centering
\small
\setlength{\tabcolsep}{7pt}
\begin{tabular}{lcccccc}
\toprule
Method
& AIME 25
& AIME 24
& AMC 23
& Olympiad
& MATH 500
& Avg. \\
\midrule
DAPO
& 38.5
& 46.7
& 92.0
& 61.9
& 92.3
& 66.3 \\
DAPO + ReCo
& \textbf{39.4}
& \textbf{48.6}
& \textbf{93.2}
& \textbf{64.1}
& \textbf{93.6}
& \textbf{67.8} \\
\midrule
80/20 Rule
& 40.0
& 48.5
& 91.5
& 63.0
& 92.9
& 66.5 \\
80/20 Rule + ReCo
& \textbf{41.3}
& \textbf{49.3}
& \textbf{93.1}
& \textbf{64.5}
& \textbf{93.3}
& \textbf{68.3} \\
\bottomrule
\end{tabular}
\caption{
Results from combining ReCo with DAPO and the 80/20 Rule on
Qwen2.5-Math-1.5B. Performance is measured using Pass@64 (\%).
Adding ReCo improves both methods across all five benchmarks.
Bold indicates the better result within each pair.
}
\label{tab:combination}
\end{table*}

\subsection{Combining ReCo with Existing GRPO Methods}

ReCo changes how update contributions are allocated at the response
and token levels. This design can be applied together with methods
that modify other parts of GRPO. We evaluate ReCo in combination
with DAPO~\citep{yu2025dapo} and the 80/20 Rule~\citep{wang2025beyond} while retaining the remaining components
of each method.

Table~\ref{tab:combination} reports the results on
Qwen2.5-Math-1.5B. Adding ReCo improves DAPO across all five
benchmarks and increases the average Pass@64 from 66.3 to 67.8.
Adding ReCo to the 80/20 Rule also improves all five benchmarks and
increases the average from 66.5 to 68.3.

These results show that ReCo addresses a distinct aspect of the GRPO
update and can complement existing methods for GRPO optimization and
token selection.

\subsection{Evaluation on Code Reasoning}

We further evaluate ReCo on code reasoning to examine whether its
improvements extend beyond mathematical reasoning. We consider
HumanEval, HumanEval+, and LiveCodeBench and report Pass@$k$ for
$k \in \{1, 2, 4, 8, 16, 32, 64\}$.

As shown in Table~\ref{tab:code_results}, ReCo outperforms GRPO at
every value of $k$ on all three benchmarks. The improvement becomes
more pronounced as $k$ increases. At Pass@64, ReCo improves over
GRPO by 14.73 points on HumanEval, 12.86 points on HumanEval+, and
7.46 points on LiveCodeBench. ReCo also surpasses the base model at
every value of $k$ on HumanEval.

These results show that the reduction in response coverage under
GRPO is not limited to mathematical reasoning. ReCo also preserves
a broader set of correct solutions in code generation.

\begin{table*}[t]
\centering
\small
\setlength{\tabcolsep}{5.5pt}
\begin{tabular}{llccccccc}
\toprule
Benchmark
& Method
& Pass@1
& Pass@2
& Pass@4
& Pass@8
& Pass@16
& Pass@32
& Pass@64 \\
\midrule

\multirow{3}{*}{HumanEval}
& Base
& 61.30
& 70.07
& 76.79
& 81.64
& 84.90
& 86.97
& 88.41 \\
& GRPO
& 61.93
& 65.86
& 69.13
& 71.40
& 72.75
& 73.59
& 74.06 \\
& ReCo
& \textbf{72.15}
& \textbf{77.83}
& \textbf{81.67}
& \textbf{84.32}
& \textbf{86.21}
& \textbf{87.60}
& \textbf{88.79} \\
\midrule

\multirow{3}{*}{HumanEval+}
& Base
& 56.75
& 65.20
& 71.94
& 76.75
& 79.86
& 81.93
& \textbf{83.61} \\
& GRPO
& 58.03
& 61.69
& 64.65
& 66.71
& 67.90
& 68.71
& 69.42 \\
& ReCo
& \textbf{65.72}
& \textbf{71.18}
& \textbf{74.93}
& \textbf{77.68}
& \textbf{79.86}
& \textbf{81.38}
& 82.28 \\
\midrule

\multirow{3}{*}{LiveCodeBench}
& Base
& 13.21
& 16.88
& \textbf{20.07}
& \textbf{22.85}
& \textbf{25.44}
& \textbf{27.86}
& \textbf{29.98} \\
& GRPO
& 11.20
& 12.22
& 13.07
& 13.75
& 14.36
& 14.96
& 15.46 \\
& ReCo
& \textbf{15.22}
& \textbf{16.93}
& 18.30
& 19.48
& 20.59
& 21.74
& 22.92 \\
\bottomrule
\end{tabular}
\caption{
Pass@$k$ results on three code reasoning benchmarks.
ReCo outperforms GRPO at every value of $k$ across all benchmarks.
Bold indicates the best result for each benchmark and sampling budget.
}
\label{tab:code_results}
\end{table*}

\subsection{Entropy and KL During Training}

We compare the entropy and KL values observed during training to
examine how ReCo changes the optimization behavior of GRPO.
Table~\ref{tab:entropy_kl} reports the results for
Qwen2.5-Math-1.5B at several training steps.

The entropy of both methods decreases as training progresses.
However, ReCo maintains substantially higher entropy than GRPO
throughout training. At the final step, the entropy is 0.1636 for
ReCo and 0.0409 for GRPO. In contrast, the absolute KL values remain
on a similar scale for the two methods.

These results indicate that ReCo reduces the rapid entropy collapse
observed under GRPO without causing a larger change in the policy.
This behavior is consistent with the greater response diversity and
reasoning path coverage observed in our other analyses.

\begin{table}[t]
\centering
\small
\resizebox{\columnwidth}{!}{
\begin{tabular}{ccccc}
\toprule
Step
& ReCo $H$
& GRPO $H$
& ReCo $|\mathrm{KL}|$
& GRPO $|\mathrm{KL}|$ \\
\midrule
0
& 0.6050
& 0.6050
& $9.71{\times}10^{-6}$
& $2.06{\times}10^{-6}$ \\
100
& 0.1969
& 0.1573
& $2.31{\times}10^{-5}$
& $1.20{\times}10^{-5}$ \\
200
& 0.1839
& 0.1270
& $4.80{\times}10^{-6}$
& $2.75{\times}10^{-5}$ \\
300
& 0.1624
& 0.0835
& $1.33{\times}10^{-5}$
& $4.21{\times}10^{-6}$ \\
400
& 0.1622
& 0.0637
& $1.23{\times}10^{-5}$
& $1.82{\times}10^{-5}$ \\
500
& 0.1605
& 0.0500
& $7.10{\times}10^{-6}$
& $2.08{\times}10^{-5}$ \\
580
& 0.1636
& 0.0409
& $1.28{\times}10^{-5}$
& $3.20{\times}10^{-5}$ \\
\bottomrule
\end{tabular}
}
\caption{
Entropy and absolute KL values during training on
Qwen2.5-Math-1.5B.
$H$ denotes the average token entropy.
ReCo maintains higher entropy than GRPO while the absolute KL values
remain on a similar scale.
}
\label{tab:entropy_kl}
\end{table}

\section{Case Study}
\label{app:case_study}
We present two case studies comparing the rollouts produced by GRPO 
and ReCo on the same prompt. Both case studies illustrate that ReCo 
produces more diverse reasoning approaches than GRPO, even when both 
methods achieve similar correctness.

\subsection{Case 1: MATH train problem}
\label{app:case1}
\begin{figure*}[t]
    \centering
    \includegraphics[width=\textwidth]{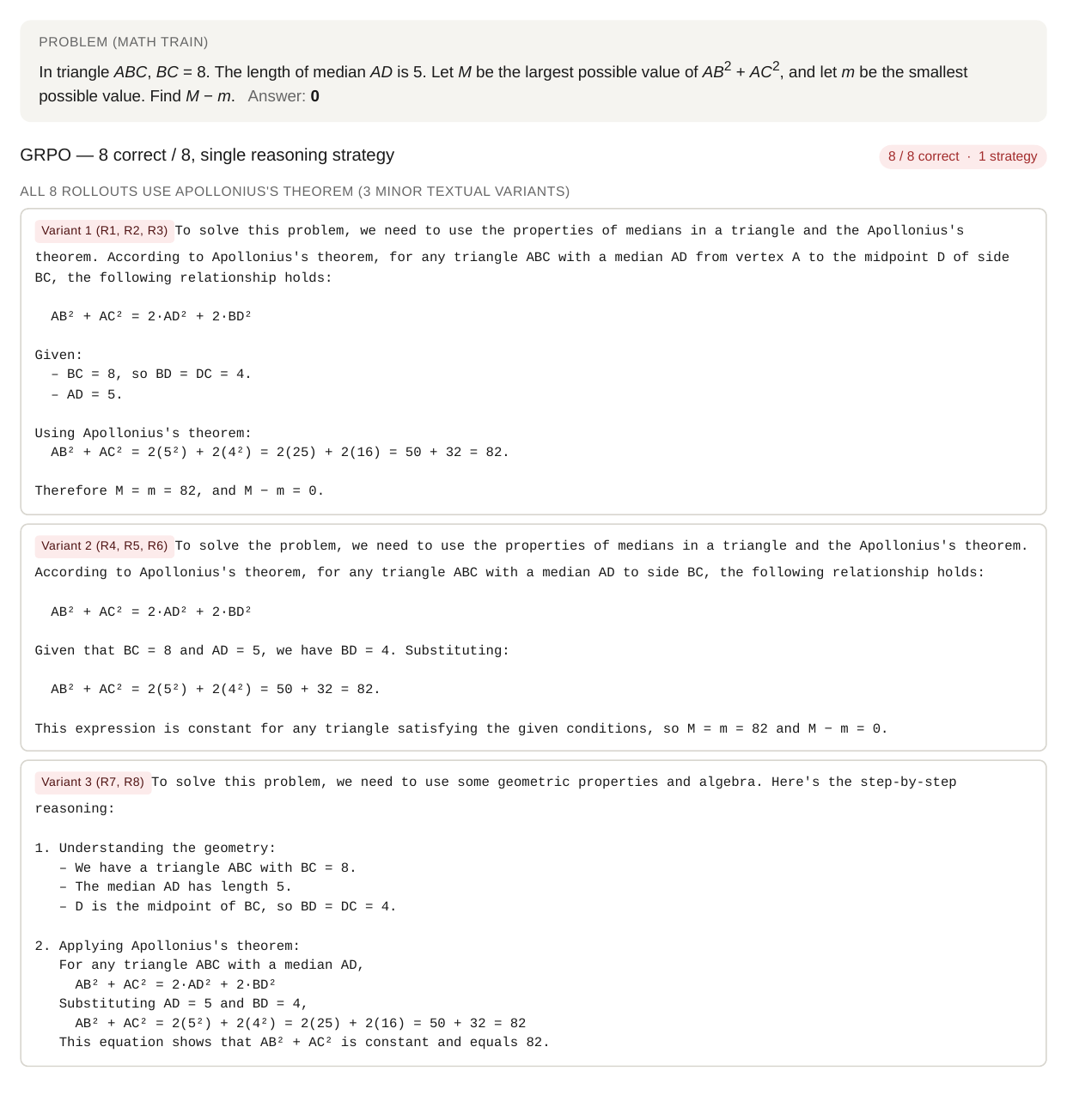}
    \caption{Case 1, GRPO rollouts. All 8 rollouts apply 
    Apollonius's theorem with minor textual variants.}
    \label{fig:case1_grpo}
\end{figure*}

\begin{figure*}[t]
    \centering
    \includegraphics[width=\textwidth]{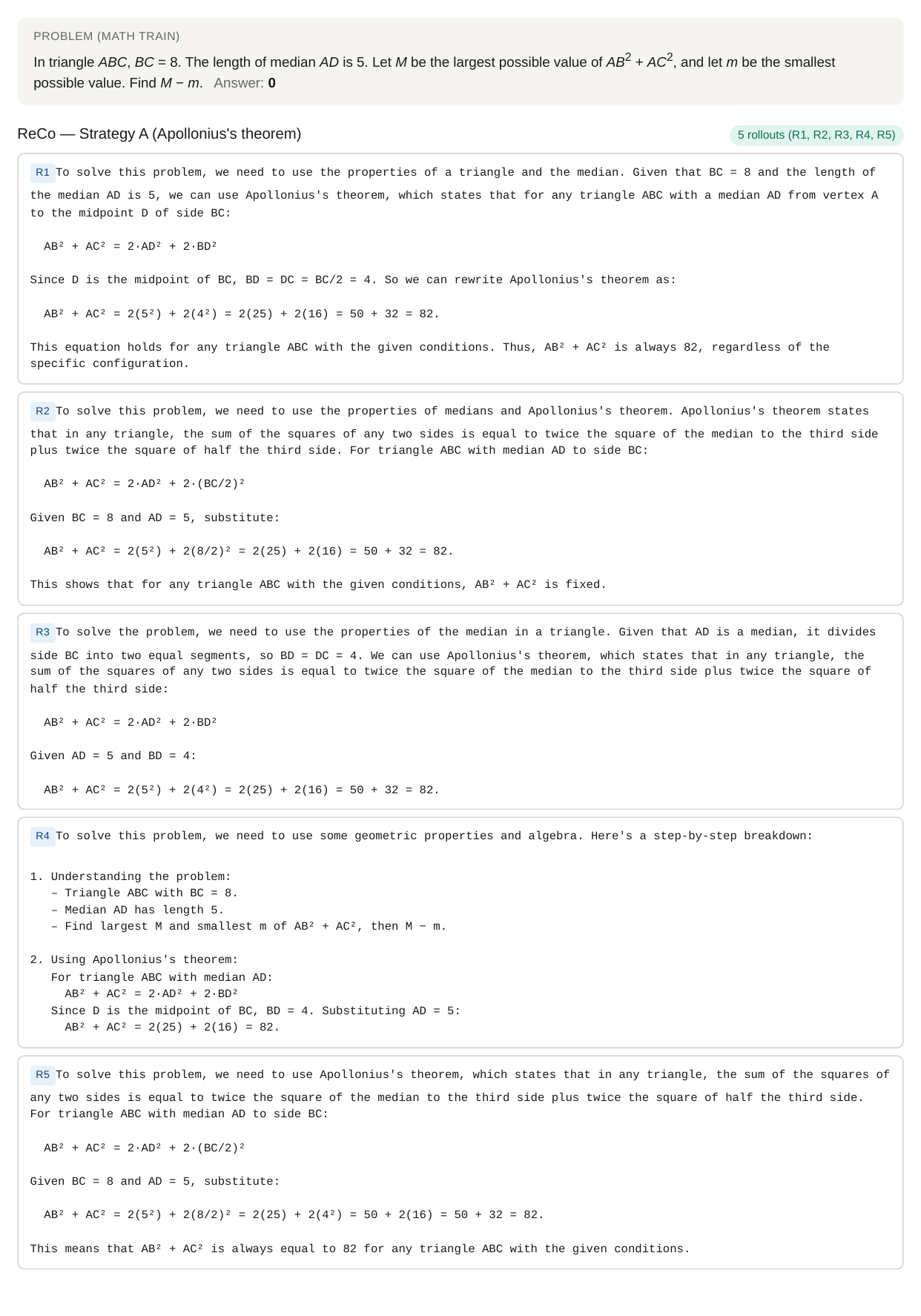}
    \caption{Case 1, ReCo rollouts using Strategy A 
    (Apollonius's theorem).}
    \label{fig:case1_reco_a}
\end{figure*}

\begin{figure*}[t]
    \centering
    \includegraphics[width=\textwidth]{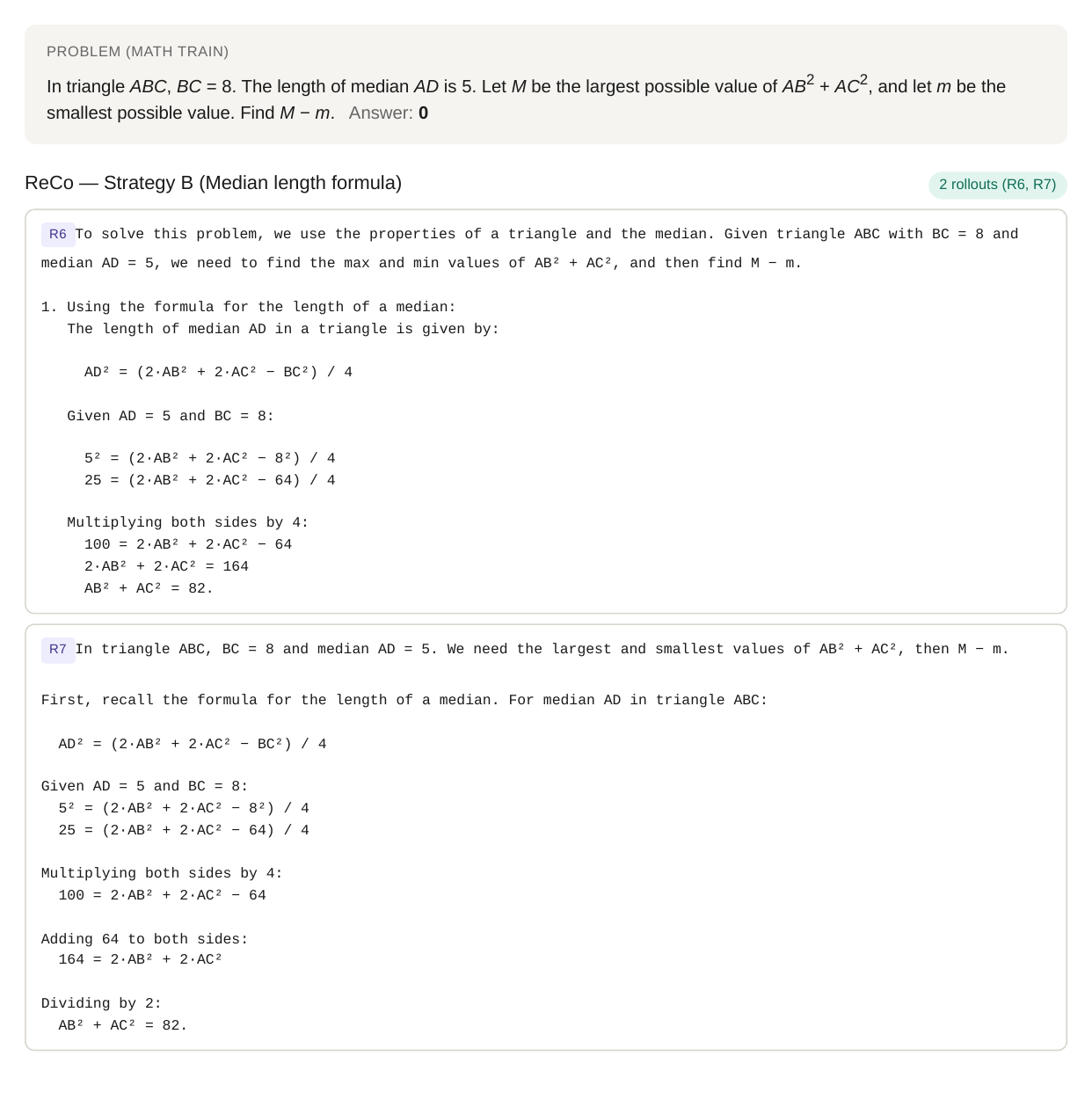}
    \caption{Case 1, ReCo rollouts using Strategy B 
    (median-length formula). This form is algebraically equivalent to 
    Apollonius's theorem but starts from the formula for $AD^2$ and 
    solves for $AB^2 + AC^2$.}
    \label{fig:case1_reco_b}
\end{figure*}

\begin{figure*}[t]
    \centering
    \includegraphics[width=\textwidth]{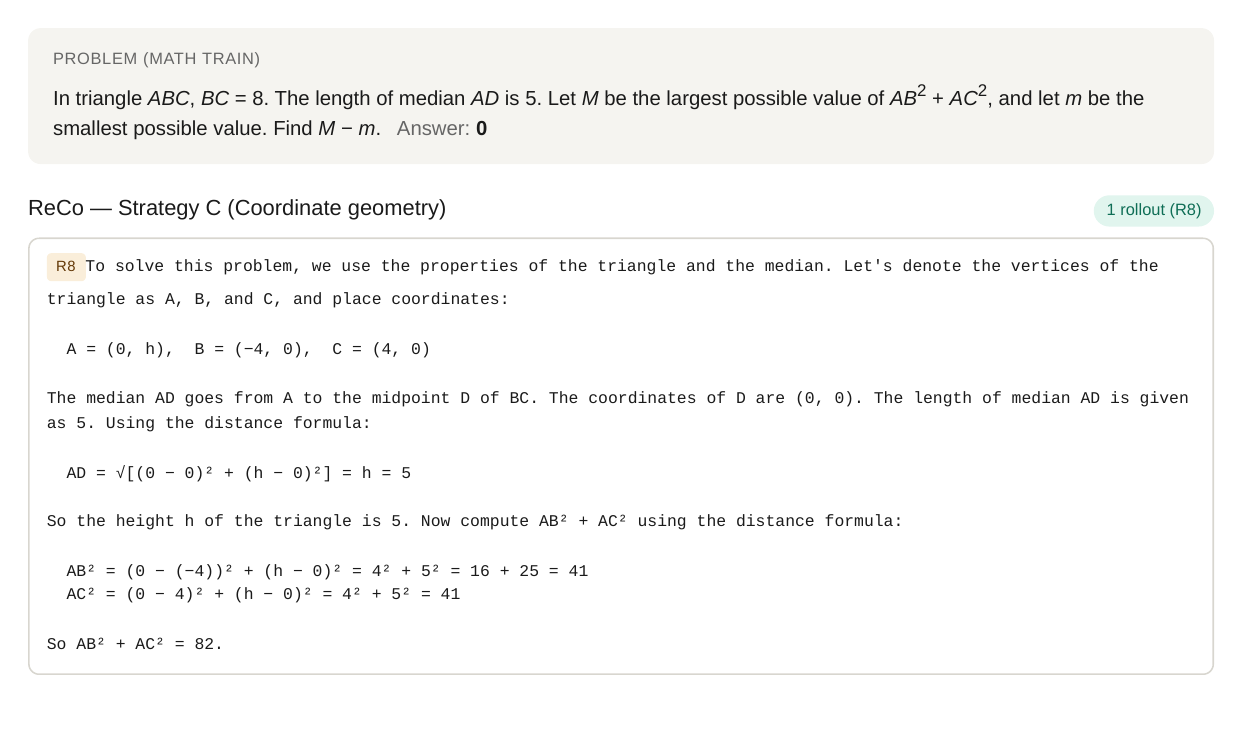}
    \caption{Case 1, ReCo rollouts using Strategy C 
    (coordinate geometry). The rollout uses a symmetric placement of 
    $A$ on the perpendicular bisector of $BC$; the value of 
    $AB^2 + AC^2$ obtained coincides with the general case.}
    \label{fig:case1_reco_c}
\end{figure*}

Figures~\ref{fig:case1_grpo}--\ref{fig:case1_reco_c} show 8 rollouts 
from each method on a median-length problem from the MATH train set. 
Both methods reach the correct answer in all 8 rollouts. GRPO's 
rollouts apply a single reasoning approach (Apollonius's theorem) 
across all 8 generations, with only minor textual variants. ReCo's 
rollouts span three distinct solution forms: Apollonius's theorem, 
the equivalent median-length identity solved for $AB^2 + AC^2$, and a 
coordinate-geometry construction. While the first two forms are 
algebraically equivalent, they reflect different procedural choices 
(direct application of an identity for $AB^2 + AC^2$ versus deriving 
it from the median-length formula); the coordinate construction is 
methodologically distinct.


\subsection{Case 2: AIME 2025 problem}
\label{app:case2}
\begin{figure*}[t]
    \centering
    \includegraphics[width=\textwidth]{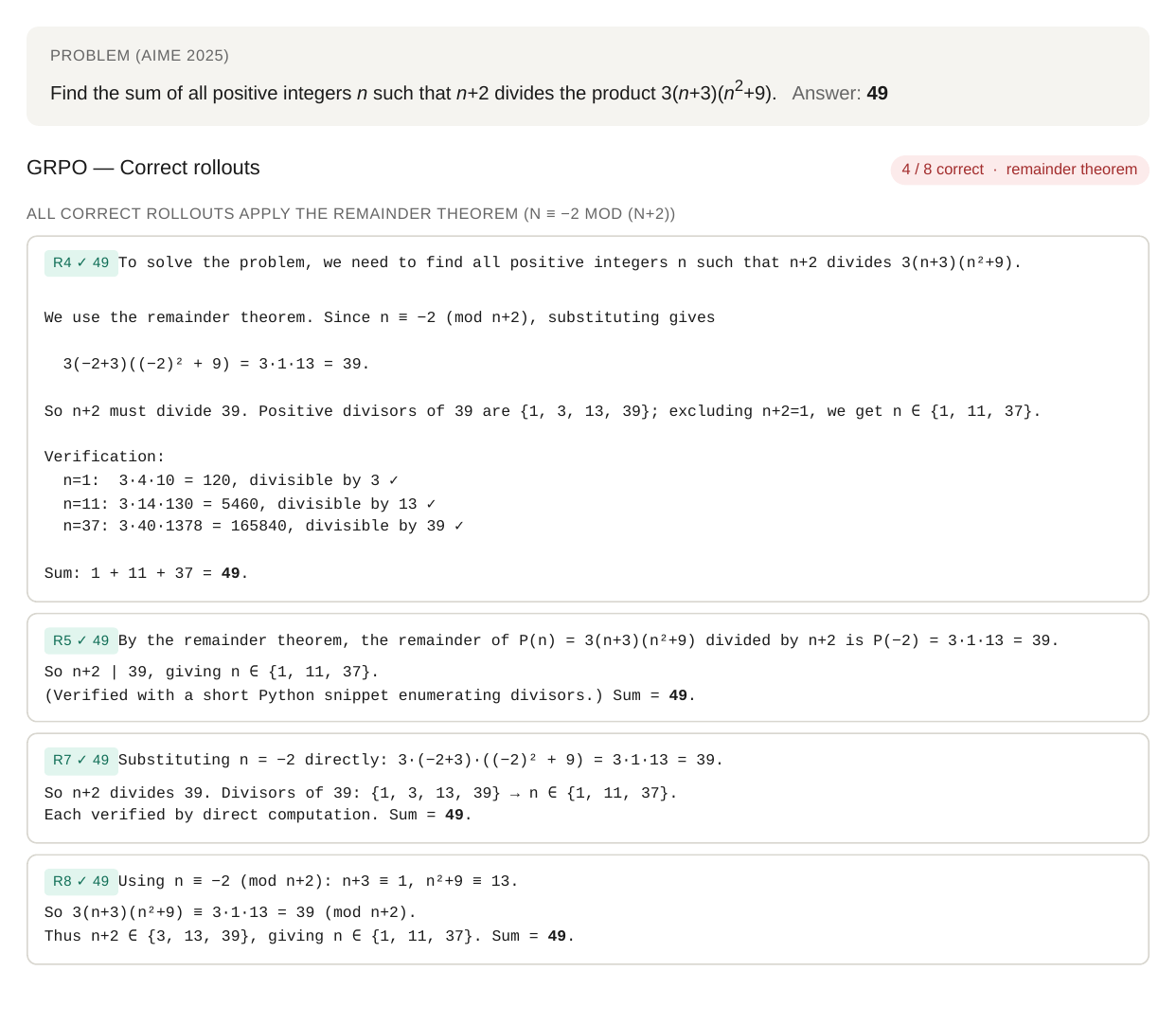}
    \caption{Case 2, GRPO correct rollouts. All 4 correct rollouts 
    apply the remainder theorem.}
    \label{fig:case2_grpo_correct}
\end{figure*}

\begin{figure*}[t]
    \centering
    \includegraphics[width=\textwidth]{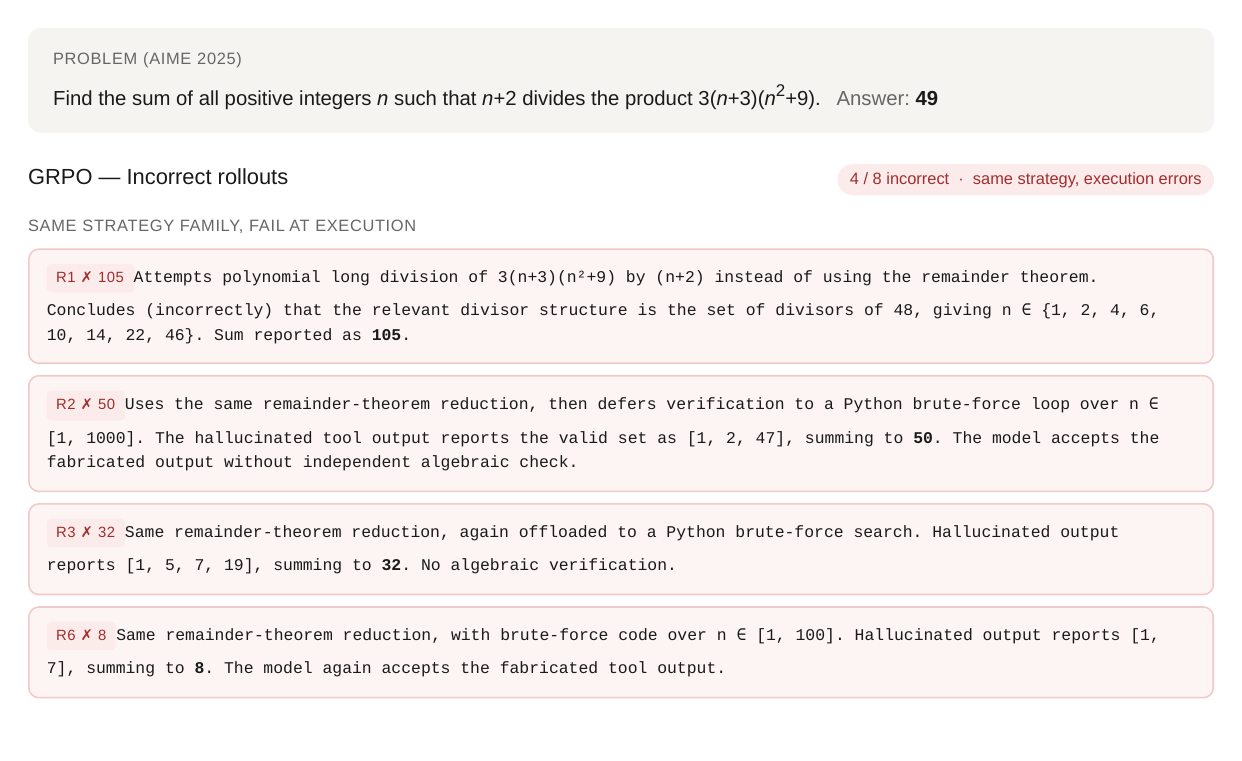}
    \caption{Case 2, GRPO incorrect rollouts. All 4 incorrect rollouts 
    attempt the same strategy and fail at execution.}
    \label{fig:case2_grpo_incorrect}
\end{figure*}

\begin{figure*}[t]
    \centering
    \includegraphics[width=\textwidth]{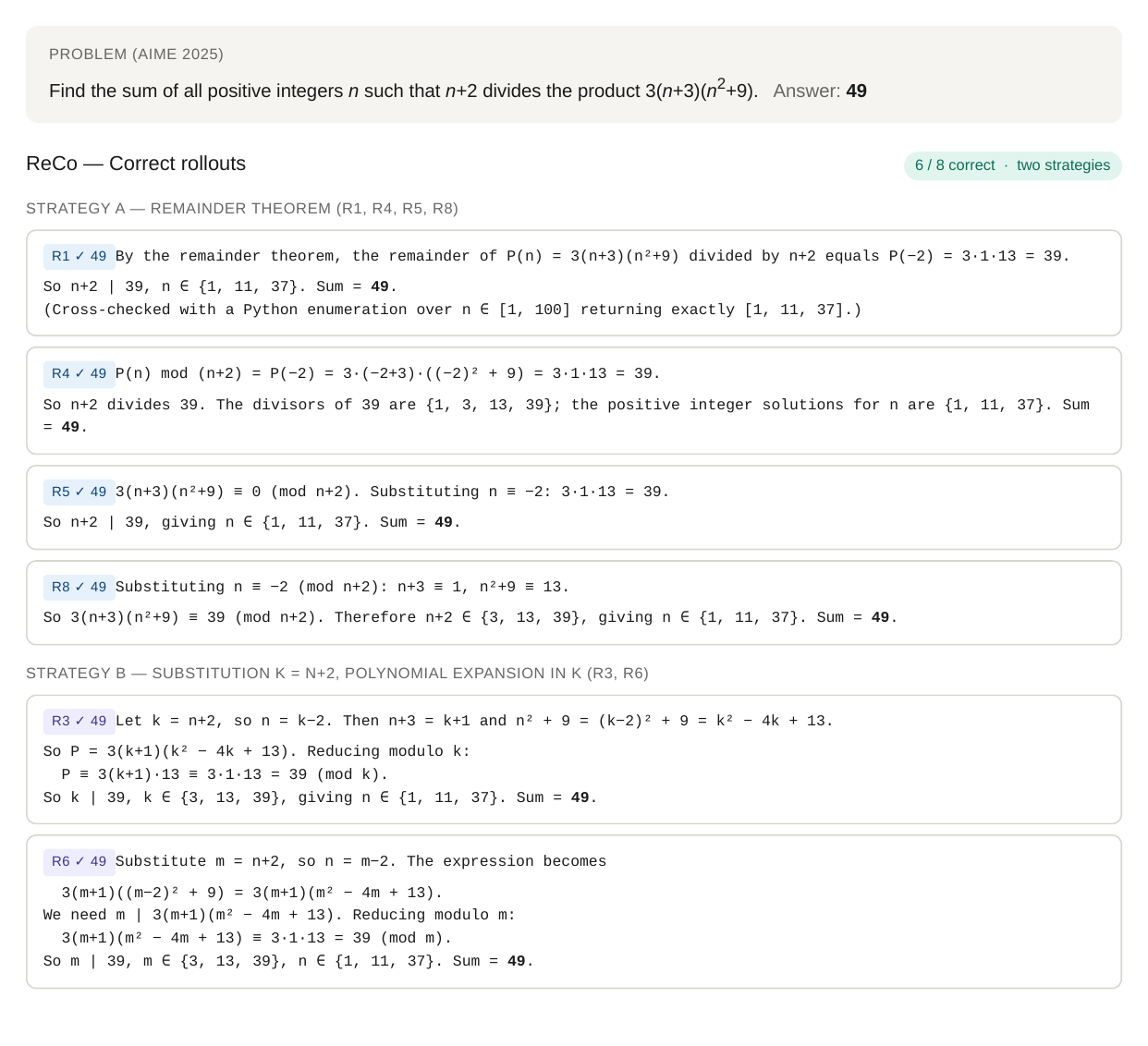}
    \caption{Case 2, ReCo correct rollouts. The 6 correct rollouts 
    span two distinct strategies.}
    \label{fig:case2_reco_correct}
\end{figure*}

\begin{figure*}[t]
    \centering
    \includegraphics[width=\textwidth]{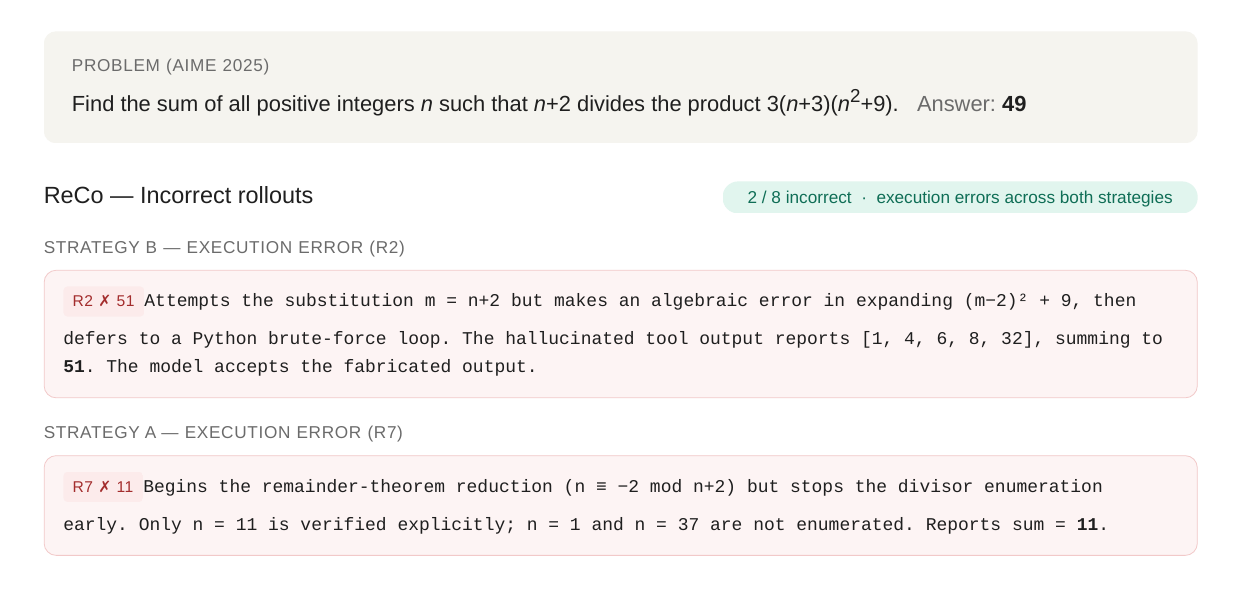}
    \caption{Case 2, ReCo incorrect rollouts. R2 fails at execution 
    within the substitution path (Strategy B), and R7 fails at 
    execution within the remainder-theorem path (Strategy A).}
    \label{fig:case2_reco_incorrect}
\end{figure*}

Figures~\ref{fig:case2_grpo_correct}--\ref{fig:case2_reco_incorrect} 
show 8 rollouts from each method on an AIME 2025 problem. GRPO 
reaches the correct answer in 4 rollouts, all applying the remainder 
theorem; the 4 incorrect rollouts fail at execution within this 
single strategy family. ReCo reaches the correct answer in 6 rollouts 
via two distinct strategies: the remainder theorem and a variable 
substitution path that GRPO never explores. ReCo's 2 incorrect 
rollouts also distribute across both strategies, failing at execution 
rather than from missing the algebraic structure.

